\documentclass[mlabstract]{jmlr}

\DeclareMathOperator{\Sym}{Sym}      
\DeclareMathOperator{\Diag}{Diag}    
\DeclareMathOperator{\rank}{rank}
\DeclareMathOperator{\range}{range}
\DeclareMathOperator{\spn}{span}     
\DeclareMathOperator{\sym}{sym}      
\DeclareMathOperator{\Lie}{Lie}
\DeclareMathOperator{\tr}{tr}

\newcommand{\rk}{\mathrm{rk}}        
 
\newcommand{\R}{\mathbb{R}}
\newcommand{\E}{\mathbb{E}}
\newcommand{\ind}{\mathbf{1}}        
\newcommand{\Wphi}{\mathcal{W}_\phi}

\newcommand{\Uc}{\mathcal{U}}

\newcommand{\Ngf}{N_{\mathrm{GF}}}

\newcommand{\Bcal}{\mathcal{B}}
\newcommand{\Btil}{\widetilde{\mathcal{B}}}

\usepackage{xcolor}

\usepackage{comment}
   
\usepackage{longtable}

\usepackage{booktabs}
\usepackage[load-configurations=version-1]{siunitx} 

\theorembodyfont{\upshape}
\theoremheaderfont{\scshape}
\theorempostheader{:}
\theoremsep{\newline}

\jmlrvolume{}
\firstpageno{1}

\jmlryear{2026}
\jmlrworkshop{Symmetry and Geometry in Neural Representations}

\title{Symmetry Breaking at the Edge of Stability \  \titlebreak via Central Flow Analysis}

   \author{\Name{Naima Elosegui Borras}\footnote{Correspondence to \Email{nelbo@dtu.dk}} \and 
   \Name{Georgios Arvanitidis} \\
   \addr Technical University of Denmark}

\begin{document}

\maketitle

\begin{abstract}
Continuous symmetries in neural networks lead to conserved quantities that are invariant under gradient flow, such as layer balancedness in linear and ReLU networks. Finite-step gradient descent need not preserve these laws, particularly at the edge of stability (EOS), where the iterates
oscillate around a smoother trajectory, which can be described by the central flow dynamics \cite{cohen2025understanding}. We study how gradient flow conservation laws evolve under the central flow description of this regime. We show that the curvature correction induces a finite-dimensional drift determined by the oscillation covariance along critical parameter directions. We thus take the first steps to analyse which conservation laws remain and are broken at the EOS according to the central flow dynamics by analysing a scalar two-layer model, two-layer matrix factorisation, and a two-layer ReLU
network, where the known conservation laws are shown to be complete \cite{marcotte2023abide}.
\end{abstract}


\section{Introduction}
\label{sec:intro}

Continuous symmetries often leave traces in optimisation dynamics. Under
gradient flow, architectural symmetries of neural networks give rise to
conserved quantities. For example, \cite{arora2019iclr-convergence} build on linear networks preserving
the matrix-valued balancedness quantity $\Psi = U^\top U - V^\top V$, while positively homogeneous networks admit corresponding neuron-wise
balancedness quantities \cite{du2018neurips-algorithmic}. \cite{marcotte2023abide} formalised this phenomenon intrinsically to the gradient flow. Writing the model through a reparametrisation $g(\theta,x)=f(\phi(\theta),x)$, they characterise gradient flow conservation laws through orthogonality to the family of admissible data-generated gradient vector fields
$\Wphi(\theta) := \range\!\left(\partial\phi(\theta)^\top\right)$, making conservation laws solely architecture-dependent and showing that the maximum number of independent laws can be computed from the dimension of the model and the Lie algebra generated by $\Wphi$, namely $\Ngf(\theta) = D-\dim\Lie(\Wphi)(\theta)$. For the shallow linear and ReLU architectures considered in this work, this machinery recovers the familiar balancedness quantities and certifies their completeness.

Finite-step gradient descent does not inherit this conservation guarantee. Symmetry-induced conservation laws are known to acquire finite-step
corrections \cite{kunin2021neural}, and gradient descent on neural networks, with learning rate $\eta$, commonly enters the EOS regime, where the largest Hessian eigenvalue (or a set of eigenvalues) approaches the stability threshold $\lambda_{\max}(H) \approx \frac{2}{\eta}$, and the iterates oscillate \cite{cohen2021iclr-gradient,damian2023iclr-selfstabilization}. \cite{cohen2025understanding} propose the \emph{central flow} as a continuous-time description of the time-averaged trajectory in this oscillatory regime.
This raises the question we study:
\begin{quote}
\emph{Which gradient flow conservation laws survive the central flow
dynamics, and which are broken by edge-of-stability oscillations?}
\end{quote}

We take the first steps to answer this question for three increasingly structured two-layer examples:
a scalar factorisation, matrix factorisation, and a ReLU network. In each
case, the gradient flow conservation laws can be represented as (where $\langle C,D \rangle$ of two matrices $C$ and $D$ denotes their Frobenius inner product)
\begin{equation} \label{eq:conservation laws} 
    \Psi_A
    =
    \langle A,\Psi\rangle,
    \qquad
    A\in\mathcal{A},
\end{equation}
a matrix-valued invariant $\Psi$, defined in Section \ref{sec:symmetries} against an architecture-dependent space of admissible test matrices \(\mathcal{A}\). 
More generally, conservation law survival at a fixed \(\theta\) is determined by the span of the symmetry-breaking directions reachable as the admissible
oscillation covariance varies, whose dimension is bounded in (\ref{eq:rank_bound}); survival along a trajectory is characterised in \eqref{eq:trajectory_survival}.

Our viewpoint is complementary to recent analyses of symmetry breaking in
the discrete dynamics. ~\cite{kunin2021neural} study finite-step
breaking through modified-equation analysis. Recent work at the edge of
stability studies balancedness and implicit bias directly in discrete
dynamics, including scalar or layer-aggregated conservation quantities
\cite{medeiros2026conservationlawbreakingedge,hofmann2026mapflowfinitestepgradient}, while
\cite{ghosh2025iclr-learning} analyse balancing in deep matrix factorisation beyond the edge of stability. Here the object of study is instead the central flow dynamics: we identify which directions in that finite-dimensional conservation-law space are broken by the EOS curvature field. We then compare gradient flow, discrete gradient descent, and the central flow to illustrate how the predicted breaking emerges along central flow trajectories.

\section{Background on symmetries and gradient flow conservation laws}
\label{sec:symmetries}

We briefly recall the framework we use following \cite{marcotte2023abide,marcotte2024keep}.
Suppose the model admits a reparametrisation
$g(\theta,x)=f(\phi(\theta),x)$,
so that, for a loss of the form \(L=F\circ\phi\),
\begin{equation}
\label{eq:loss_GF}
    \nabla L(\theta)
    =
    \partial\phi(\theta)^\top \nabla F(\phi(\theta))
    \in
    \Wphi(\theta),
    \qquad
    \Wphi(\theta)
    :=
    \range\!\left(\partial\phi(\theta)^\top\right).
\end{equation}
A \(C^1\) function \(h\) is therefore conserved by gradient flow whenever $\nabla h(\theta)\perp \Wphi(\theta)$, since $\dot h = -\langle \nabla h,\nabla L\rangle = 0$. The maximal number of locally independent laws, under the assumption that the loss spans the output space and that $\dim\Lie(\Wphi)$ is locally constant, is $\Ngf(\theta) = D-\dim\Lie(\Wphi)(\theta)$.

The conservation laws used below arise from continuous invariances of the loss \eqref{eq:loss_GF}. Let $T : \mathbb{R} \times  \Omega\rightarrow \mathbb{R}^D$ \ be a one parameter transformation on an open subset $\Omega$ of the model parameter space satisfying $T(0,\theta)=\theta, \phi(T(\epsilon,\theta))=\phi(\theta)$. Its infinitesimal generator
$
    \Delta(\theta)
    :=
    \left(
    \frac{\mathrm{d}}{\mathrm{d}\epsilon}
    T(\epsilon,\theta)
    \right|_{\epsilon=0}
$
satisfies
$\partial\phi(\theta)\Delta(\theta) = 0$,
and hence lies in \(\Wphi(\theta)^\perp\), leaving the loss invariant. If, in addition, \(\Delta (\theta)=\nabla h (\theta)\) for some scalar potential \(h\), then \(h\) is a
gradient flow conservation law. Importantly, differentiating an invariance
only implies that its generator lies in \(\ker\partial\phi (\theta)\); it does not
require every element of this kernel to arise from the particular symmetry
family considered below.

\paragraph{Two-layer matrix factorisation.}
Let
$
    \theta=(U,V),
    U\in\R^{n\times r},
    V\in\R^{m\times r},
    \phi(U,V)=UV^\top
$.
For any \(A\in\R^{r\times r}\), the transformation $
    T_A(\epsilon,U,V)
    =
    \left(
        Ue^{\epsilon A},
        Ve^{-\epsilon A^\top}
    \right)
$
leaves \(UV^\top\) unchanged, with generator
 $\Delta_A(U,V) :=\frac{d}{d\epsilon}T_A(\epsilon,U,V)|_{\epsilon = 0}
    =
    \left(
        UA,
        -VA^\top
    \right).
$
This vector field is a Euclidean gradient field precisely when
\(A\in\Sym(r)\). For such \(A\), define
$
    \Psi
    :=
    U^\top U-V^\top V
    \in\Sym(r),
    \Psi_A
    :=
    \langle A,\Psi\rangle
    =
    \tr(A\Psi).
$
Direct differentiation gives $\nabla\Psi_A
    =
    \left(
        2UA,
        -2VA
    \right)
    =
    2\Delta_A$. Thus the admissible test space for matrix factorisation gradient flow laws is
$
    \mathcal A_{\mathrm{MF}}
    =
    \Sym(r)
$.
If $(U;V)$ has full rank $\rk$,
a complete set of locally independent laws has dimension $
    \Ngf
    =
    \frac{\rk(2r+1-\rk)}{2},
$ \cite{marcotte2023abide}.
\paragraph{Two-layer ReLU.}
Consider
$
    g(\theta,x)
    =
    U\sigma(V^\top x)
    =
    \sum_{j=1}^r
    u_j\,\sigma(v_j^\top x),
$
with \(\sigma(z)=\max\{z,0\}\). Positive homogeneity gives the
neuron-wise invariance
$
    (u_j,v_j)
    \longmapsto
    \left(
        e^{\epsilon\lambda_j}u_j,
        e^{-\epsilon\lambda_j}v_j
    \right)
$. Equivalently, within the matrix family above the admissible continuous
generators are, generically, $
    A\in\Diag(r)$.
The corresponding independent conservation laws are
$
    h_j
    =
    \|u_j\|_2^2-\|v_j\|_2^2,
    j=1,\ldots,r
$,
or, for \(A=\text{diag}(\lambda_1,\ldots,\lambda_r)\),
$
    \Psi_A
    =
    \sum_{j=1}^r\lambda_j h_j.
$
Hence
$
    \mathcal A_{\mathrm{ReLU}}
    =
    \Diag(r), \text{and} \ 
    \Ngf=r.
$
These laws form a complete set of independent gradient flow conservation laws for the two-layer ReLU architecture \cite{marcotte2023abide}.

The scalar model is the special case \(r=1\). Thus all three examples admit
the common representation
in Equation \ref{eq:conservation laws}
but the admissible test space is architecture-dependent:
$
    \mathcal A
    =
    \R,
    \Sym(r),
    \Diag(r)
$, for the scalar, matrix-factorisation, and ReLU examples, respectively.
Section~\ref{sec:breaking} shows that this distinction determines which
component of the central flow perturbation can break the corresponding laws.

\section{Symmetry breaking under the central flow}
\label{sec:breaking}

We now ask how the gradient flow conservation laws of
Section~\ref{sec:symmetries} evolve under the central flow. \cite{cohen2025understanding} model the time-averaged trajectory of gradient
descent at the edge of stability by
\begin{equation}
    \dot{\theta}
    =
    -\eta\nabla L(\theta)
    -
    \frac{\eta}{2}
    \nabla_\theta
    \langle H(\theta),\Sigma\rangle,
    \qquad
    H(\theta):=\nabla^2L(\theta).
    \label{eq:central_flow}
\end{equation}
Here \(\Sigma\succeq0\) models the covariance of the fast oscillations.
At the edge of stability its range is contained in the critical eigenspace
\begin{equation}
    \Uc(\theta)
    :=
    \ker\left(
        H(\theta)-\frac{2}{\eta}I
    \right),
    \qquad
    k:=\dim\Uc(\theta),
    \label{eq:critical_space}
\end{equation}
while the solution to the \emph{semidefinite complementarity problem} (SDCP) selects the realised
\(\Sigma(t)\), pinning the active eigenvalues at
\(2/\eta\). Away from the edge of stability, the selected covariance is
\(\Sigma=0\), and~\eqref{eq:central_flow} reduces to gradient flow. Writing an oscillatory displacement as \(\delta\), with \(\E[\delta]=0\) and \(\Sigma=\E[\delta\delta^\top]\), the curvature contraction has the equivalent form
\begin{equation}
    \langle H,\Sigma\rangle
    =
    \tr(H^\top\Sigma)
    =
    \E[\delta^\top H\delta].
    \label{eq:hessian_covariance}
\end{equation}
Thus \(\Sigma\) records where the finite-step dynamics oscillate, while
\(\langle H,\Sigma\rangle\) measures the curvature encountered by those
oscillations. 

\paragraph{The central flow conservation law drift.}
Let \(h\) be a gradient flow conservation law. Since
\(\langle\nabla h,\nabla L\rangle=0\), contracting
\eqref{eq:central_flow} with \(\nabla h\) eliminates the gradient flow
field:
\begin{equation}
    \dot h
    =
    -\frac{\eta}{2}
    D_{\nabla h}
    \langle H(\theta),\Sigma\rangle .
    \label{eq:general_defect}
\end{equation}
The directional derivative in~\eqref{eq:general_defect} is taken with
\(\Sigma\) held fixed. It is therefore not the total time derivative of
\(\langle H(\theta(t)),\Sigma(t)\rangle\), which would contain an
additional term involving \(\dot{\Sigma}\).

For the laws of Section~\ref{sec:symmetries},
\(\nabla\Psi_A=2\Delta_A\), so
\begin{equation}
    \dot{\Psi}_A
    =
    -\eta
    D_{\Delta_A}
    \langle H(\theta),\Sigma\rangle .
    \label{eq:psi_defect}
\end{equation}

\paragraph{Collapse to the factor dimension.}
Write \(L=F\circ\phi\). The Hessian decomposes as
\begin{equation}
    H
    =
    H_{\mathrm p}+H_{\mathrm c},
    \qquad
    H_{\mathrm p}
    =
    \partial\phi^\top
    (\nabla^2F)
    \partial\phi,
    \qquad
    H_{\mathrm c}
    =
    \sum_a
    (\partial_aF)\nabla^2\phi_a.
    \label{eq:hessian_split}
\end{equation}
For the squared losses considered below, \(\nabla^2F=I\) for the MF and scalar cases as well as for the ReLU case as considered in Appendix \ref{app:relu} we get
\begin{equation}
    \langle H_{\mathrm p},\Sigma\rangle 
    =
    \E
    \left[
        \|\partial\phi(\theta)\delta\|^2
    \right].
    \label{eq:pullback_contraction}
\end{equation}
In our two-layer examples, \(H_{\mathrm c}\) is invariant along the
symmetry generator, so
$
    D_{\Delta_A}
    \langle H_{\mathrm c},\Sigma\rangle
    =
    0.
    \label{eq:coordinate_inert}
$
For ReLU, this statement and the explicit Hessian calculation below are
local to a region in which the activation pattern is fixed. Because \(\partial\phi(\theta)\delta\) is bilinear in the two factors,
\eqref{eq:pullback_contraction} reduces to \(r\times r\) contractions.
Writing
\begin{equation}
    P_U:=U^\top U,
    \qquad
    P_V:=V^\top V,
    \qquad
    R_U:=\E[\delta_U^\top\delta_U],
    \qquad
    R_V:=\E[\delta_V^\top\delta_V],
    \label{eq:gram_definitions}
\end{equation}
differentiation along \(\Delta_A\) yields a linear functional of \(A\).
For all three examples this functional is in the form of the raw symmetry-breaking matrix
\begin{equation}
    \Btil
    =
    \left(
        P_U R_V+R_VP_U
    \right)
    -
    \left(
        P_VR_U+R_UP_V
    \right)
    +
    2\left(N-N'\right),
    \label{eq:raw_breaking_matrix}
\end{equation}
where $N := \E[
        (\delta_V^\top V)
        (\delta_U^\top U)
    ],
    N'
    :=
    \E[
        (\delta_U^\top U)
        (\delta_V^\top V)
    ]$ denote the two orderings of the cross contraction.
Their explicit forms, and the corresponding ReLU substitutions, are given
in the examples in Apps.~\ref{app:scalar}--\ref{app:relu}. Importantly, every term in~\eqref{eq:raw_breaking_matrix}
is linear in \(\Sigma\). The raw matrix \(\Btil\) need not belong to the architecture's admissible
test space \(\mathcal A\). We therefore define the canonical symmetry-breaking matrix by
$
    \Bcal(\theta,\Sigma)
    :=
    P_{\mathcal A}\Btil(\theta,\Sigma)
    \in\mathcal A,
    \label{eq:breaking_matrix}
$
where \(P_{\mathcal A}\) is the orthogonal projection onto
\(\mathcal A\). Equation~\eqref{eq:psi_defect} then becomes
\begin{equation}
    \boxed{
    \dot{\Psi}_A
    =
    -\eta
    \langle A,\Bcal(\theta,\Sigma)\rangle ,
    \qquad
    A\in\mathcal A.
    }
    \label{eq:breaking_law}
\end{equation}

Criterion for survival of the conservation laws is given in Appendix \ref{app:survival}. The three examples differ in their admissible test space and in the low-dimensional contractions entering the raw breaking matrix \(\Btil\).
Table~\ref{tab:examples} summarises the resulting conservation laws and observable symmetry-breaking matrices; full derivations and empirical results are given in Apps.~\ref{app:scalar}--\ref{app:relu}.

\section{Conclusions and outlook}
In this work we show how the central flow dynamics can break conservation laws that hold on gradient flows in scalar networks, matrix factorisation and two-layer ReLU networks. We identify a symmetry-breaking drift matrix, laying the ground for future analysis of more complex architectures and EOS conservation laws. The limitations of this work, that future work will address, include actual counting and completeness arguments, to identify the number of independent conservation laws that survive under central flow dynamics. Moreover, a formal distinction of uniform (for all admissible $\Sigma(t)$) and trajectory-dependent laws should be elaborated, to answer the question of whether new trajectory-dependent laws emerge at the EOS under the central flow.


\bibliography{pmlr-sample}
\newpage
\appendix
\section{Survival of gradient flow conservation laws}\label{app:survival}

\begin{table}[h]
    \centering
    \small
    \begin{tabular}{c|c|c|c}
        model
        & \(\mathcal A\)
        & gradient flow laws
        & \(\Bcal\)
        \\
        \hline
        scalar
        & \(\R\)
        & \(h=u^2-v^2\)
        & \(2(u^2\sigma_{vv}-v^2\sigma_{uu})\)
        \\
        matrix factorisation
        & \(\Sym(r)\)
        & \(\Psi_A=\tr(A\Psi)\)
        & \(\sym(\Btil)\)
        \\
        two-layer ReLU
        & \(\Diag(r)\)
        & \(h_j=\|u_j\|^2-\|v_j\|^2\)
        & \(P_{\Diag(r)}\Btil_{\rm ReLU}\)
    \end{tabular}
    \caption{
    The three instances of the symmetry-breaking construction.
    Here \(\Psi=U^\top U-V^\top V\).
    The raw matrix \(\Btil\) is projected onto the architecture-dependent
    admissible test space before determining survival.
    }
    \label{tab:examples}
\end{table}
For the realised central flow covariance \(\Sigma(t)\), define
\begin{equation}
    \Bcal(t)
    :=
    \Bcal(\theta(t),\Sigma(t)).
\end{equation}
Equation~\eqref{eq:breaking_law} immediately gives
\begin{equation}
    \Psi_A
    \text{ is conserved along the trajectory}
    \quad\Longleftrightarrow\quad
    A
    \perp
    \spn_{t}
    \left\{
        \Bcal(t)
    \right\}.
    \label{eq:trajectory_survival}
\end{equation}
Hence the central flow breaks conservation laws geometrically: the realised matrices \(\Bcal(t)\) span the broken directions, and their orthogonal complement gives the surviving constant-$A$ conservation laws along this trajectory. It is also useful to characterise all breaking directions that can be generated by covariance perturbations supported on the current critical
eigenspace. Since \(\Btil\) is linear in \(\Sigma\), we define
\begin{equation}
    M_\theta:
    \Sym(\Uc(\theta))
    \longrightarrow
    \mathcal A,
    \qquad
    M_\theta(\Sigma)
    :=
    P_{\mathcal A}
    \Btil(\theta,\Sigma).
    \label{eq:breaking_map}
\end{equation}
This is a linear extension over the span of supported covariance matrices; the actual central flow selects a \(\Sigma_{\mathrm{CF}}(\theta)\) through its SDCP solution.
Since
\begin{equation}
    \dim\Sym(\Uc(\theta))
    =
    \frac{k(k+1)}{2},
\end{equation}
the potential breaking space obeys
\begin{equation}
    \rank M_\theta
    \leq
    \min
    \left\{
        \frac{k(k+1)}{2},
        \dim\mathcal A
    \right\}.
    \label{eq:rank_bound}
\end{equation}
In particular, when \(k=1\), all covariance perturbations supported on the
critical eigenspace produce at most one breaking direction. For a single
realised \(\Sigma(t)\), independently of \(k\), the instantaneous drift is
always the single functional
\(A\mapsto\langle A,\Bcal(t)\rangle \); larger \(k\) matters because it
allows a larger set of breaking directions to be reached as the covariance
changes.

\section{Scalar factorisation}
\label{app:scalar}

Consider
\( \phi(u,v)=uv \) and
\( L(u,v)=\frac12(uv-y)^2 \).
Writing
\[
    \rho:=uv-y,
    \qquad
    h:=u^2-v^2,
\]
we have
\[
    \nabla L=\rho(v,u)^\top,
    \qquad
    \nabla h=(2u,-2v)^\top,
    \qquad
    \langle\nabla h,\nabla L\rangle=0.
\]
Hence \(h\) is, up to functions of \(h\), the unique gradient-flow conservation law in this
\(r=1\) case.

\paragraph{Central-flow drift.}
The Hessian and covariance are
\[
    H=
    \begin{pmatrix}
        v^2 & 2uv-y\\
        2uv-y & u^2
    \end{pmatrix},
    \qquad
    \Sigma=
    \begin{pmatrix}
        \sigma_{uu} & \sigma_{uv}\\
        \sigma_{uv} & \sigma_{vv}
    \end{pmatrix}.
\]
Holding \(\Sigma\) fixed,
\begin{align*}
    \langle H,\Sigma\rangle 
    &=
    v^2\sigma_{uu}
    +u^2\sigma_{vv}
    +2(2uv-y)\sigma_{uv},\\
    \nabla_\theta\langle H,\Sigma\rangle 
    &=
    \bigl(
        2u\sigma_{vv}+4v\sigma_{uv},
        \;
        2v\sigma_{uu}+4u\sigma_{uv}
    \bigr)^\top .
\end{align*}
Contracting with \(\nabla h\), the cross-covariance cancels:
\[
    D_{\nabla h}\langle H,\Sigma\rangle 
    =
    4(u^2\sigma_{vv}-v^2\sigma_{uu}).
\]
Therefore
\begin{equation}
    \boxed{
    \dot h
    =
    -2\eta
    \left(
        u^2\sigma_{vv}-v^2\sigma_{uu}
    \right)
    }.
    \label{eq:scalar_drift_app}
\end{equation}
Thus the scalar symmetry-breaking quantity is
\[
    \Bcal
    =
    2(u^2\sigma_{vv}-v^2\sigma_{uu}).
\]

\paragraph{One critical direction.}
Suppose the top Hessian eigenvalue is simple and the active critical
eigenspace is one-dimensional. Then
\(\Sigma=\sigma^2qq^\top\), where \(q\) is the unit top eigenvector, $\sigma^2$ is the variance of the oscillations along $q$, and
\[
    \langle H,\Sigma\rangle 
    =
    \sigma^2\lambda_{\max}(H)
\]
and
\[
    D_{\nabla h}\langle H,\Sigma\rangle 
    =
    \sigma^2
    \langle\nabla h,\nabla\lambda_{\max}\rangle.
\]

Let
\[
    b:=2uv-y,
    \qquad
    R:=
    \sqrt{\frac{h^2}{4}+b^2}.
\]
The largest eigenvalue is
\[
    \lambda_{\max}
    =
    \frac{u^2+v^2}{2}+R.
\]
Direct differentiation gives
\begin{align*}
    \partial_u\lambda_{\max}
    &=
    u+\frac{\frac12 uh+2bv}{R},\\
    \partial_v\lambda_{\max}
    &=
    v+\frac{-\frac12 vh+2bu}{R},
\end{align*}
and therefore
\[
    \langle\nabla h,\nabla\lambda_{\max}\rangle
    =
    h\left(
        2+\frac{u^2+v^2}{R}
    \right).
\]
Consequently,
\begin{equation}
    \boxed{
    \dot h
    =
    -\frac{\eta\sigma^2}{2}
    h
    \left(
        2+\frac{u^2+v^2}{R}
    \right)
    }.
    \label{eq:scalar_rankone_app}
\end{equation}
For \(R>0\) and \(\sigma^2>0\),
\[
    h\dot h\leq0,
\]
with equality if and only if \(h=0\). This is an instantaneous statement
during the active oscillatory phase; it does not imply that \(h(t)\)
reaches zero before \(\Sigma\) reaches zero.

\paragraph{Stable terminal values.}
On the solution manifold \(uv=y\),
\[
    \lambda_{\max}
    =
    u^2+v^2
    =
    \sqrt{h^2+4y^2}.
\]
Any stable solution-manifold endpoint satisfies
\(\eta\lambda_{\max}\leq2\), and hence
\begin{equation}
    |h_\infty|
    \leq
    2\sqrt{\frac{1}{\eta^2}-y^2},
    \qquad
    \frac{1}{\eta^2}\geq y^2.
    \label{eq:scalar_terminal_bound_app}
\end{equation}
Thus the EOS dynamics can reduce the balancedness gap and then freeze at a
generically nonzero value once the oscillatory phase terminates.

\begin{figure}[htbp]
    \centering
    \includegraphics[width=1.0\linewidth]{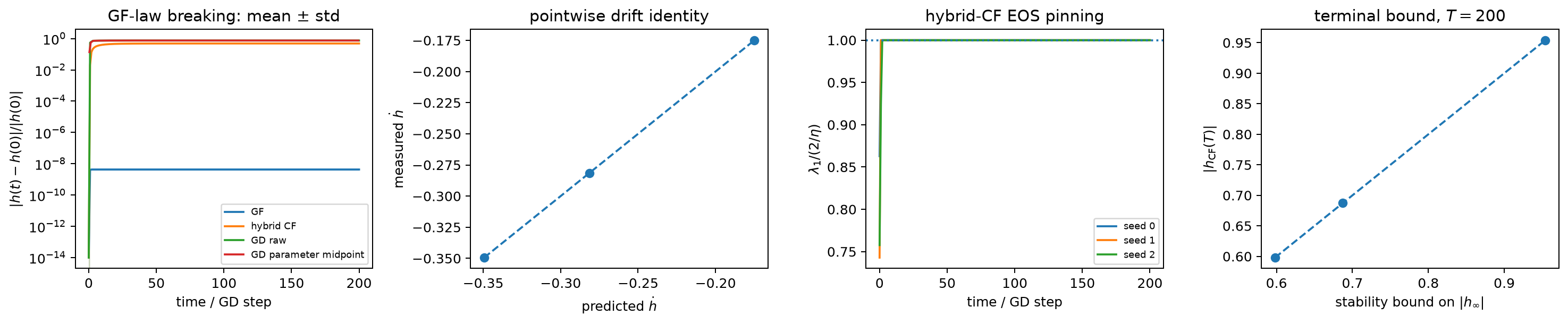}
    \caption{
\textbf{Scalar factorisation: central-flow breaking of the gradient-flow
balancedness law.}
Results are shown over three random initialisations.
\textbf{Left:} relative change of the gradient-flow invariant
\(h=u^2-v^2\). Gradient flow preserves \(h\) up to numerical integration
error, whereas finite-step GD and the hybrid central flow produce an
\(O(1)\) change. The GD parameter midpoint,
\((\theta_k+\theta_{k+1})/2\), removes the dominant period-two
oscillation before evaluating \(h\).
\textbf{Centre-left:} pointwise test of the central-flow drift at the first
EOS contact. For each seed, the directly measured
\(\dot h=\langle\nabla h,\dot\theta_{\rm CF}\rangle\) agrees with the
closed-form prediction of Eq.~\eqref{eq:scalar_drift_app},
with the dashed line denoting equality.
\textbf{Centre-right:} normalised sharpness along hybrid central flow.
Each trajectory follows ordinary gradient flow before EOS and, once the
constraint becomes active, remains pinned at
\(\lambda_{\max}(H)=2/\eta\).
\textbf{Right:} long-time value of \(|h|\) at \(T=200\) against the
stability bound
\(2\sqrt{\eta^{-2}-y^2}\).
The three runs lie on the identity line, showing empirically that the
bound is saturated in these instances.
}
    \label{fig:scalar}
\end{figure}

\paragraph{Empirical analysis.}
Figure~\ref{fig:scalar} isolates the symmetry-breaking mechanism. Gradient flow remains on the level set
\(h=h_0\), while both gradient descent (GD) and the central flow (CF) leave that level set once EOS effects become active. When reaching the EOS dynamics, the independently measured \(\dot h\) agrees with the analytic prediction. Since \(h>0\) in these runs and \(\dot h<0\), the curvature correction moves the two factors toward balance, consistent with
the sign relation \(h\dot h\leq0\).

\section{Two-layer matrix factorisation}
\label{app:mf}

Let
\[
    U\in\R^{n\times r},
    \qquad
    V\in\R^{m\times r},
    \qquad
    \phi(U,V)=UV^\top,
\]
with squared loss
\[
    L(U,V)
    =
    \frac12\|UV^\top-Y\|^2,
    \qquad
    G:=UV^\top-Y
\]
where $\|\cdot\|$ denotes the Frobenius norm. The gradient-flow invariant is
\[
    \Psi=U^\top U-V^\top V,
    \qquad
    \Psi_A=\tr(A\Psi),
    \qquad
    A\in\Sym(r),
\]
with \(\nabla\Psi_A=2\Delta_A\) and
\(\Delta_A=(UA,-VA)\).

\paragraph{Hessian contraction.}
For an oscillatory perturbation
\(\delta=(\delta_U,\delta_V)\),
\[
    \partial\phi(\theta)\delta
    =
    \delta_UV^\top+U\delta_V^\top.
\]
Since \(\phi\) is bilinear, the Hessian quadratic form splits as
\[
    \delta^\top H\delta
    =
    \underbrace{
    \|\delta_UV^\top+U\delta_V^\top\|^2
    }_{H_{\mathrm p}}
    +
    \underbrace{
    2\langle G,\delta_U\delta_V^\top\rangle 
    }_{H_{\mathrm c}}.
\]
Along the symmetry generator,
\[
    D_{\Delta_A}(UV^\top)
    =
    UAV^\top-UAV^\top
    =
    0,
\]
so \(G\) is unchanged and therefore
\[
    D_{\Delta_A}
    \langle H_{\mathrm c},\Sigma\rangle 
    =
    0.
\]
Only the pullback term contributes to symmetry breaking.

Define
\[
    P_U:=U^\top U,
    \qquad
    P_V:=V^\top V,
    \qquad
    R_U:=\E[\delta_U^\top\delta_U],
    \qquad
    R_V:=\E[\delta_V^\top\delta_V].
\]
Expanding the pullback quadratic form,
\begin{align*}
    \|\delta_UV^\top+U\delta_V^\top\|^2
    &=
    \tr(P_V\delta_U^\top\delta_U)
    +
    \tr(P_U\delta_V^\top\delta_V)\\
    &\quad
    +
    2\tr(V\delta_U^\top U\delta_V^\top).
\end{align*}
Hence
\[
    \langle H_{\mathrm p},\Sigma\rangle 
    =
    \tr(P_VR_U)
    +
    \tr(P_UR_V)
    +
    2\tr(VM),
    \qquad
    M:=\E[\delta_U^\top U\delta_V^\top].
\]

\paragraph{Differentiation along the symmetry.}
Holding \(\Sigma\) fixed gives
\[
    \dot P_U=AP_U+P_UA,
    \qquad
    \dot P_V=-(AP_V+P_VA).
\]
Therefore
\[
    \frac{\mathrm d}{\mathrm dt}
    \bigl[
        \tr(P_UR_V)+\tr(P_VR_U)
    \bigr]
    =
    \tr\!\left(
        A[
            P_UR_V+R_VP_U
            -
            P_VR_U-R_UP_V
        ]
    \right).
\]

For the cross term,
\[
    \dot M
    =
    \E[\delta_U^\top UA\delta_V^\top],
\]
so
\[
    \frac{\mathrm d}{\mathrm dt}\,2\tr(VM)
    =
    2\tr(\dot VM)+2\tr(V\dot M).
\]
Define
\[
    N
    :=
    \E[
        (\delta_V^\top V)
        (\delta_U^\top U)
    ],
    \qquad
    N'
    :=
    \E[
        (\delta_U^\top U)
        (\delta_V^\top V)
    ].
\]
Then
\[
    \tr(\dot VM)
    =
    -\tr(AN'),
    \qquad
    \tr(V\dot M)
    =
    \tr(AN),
\]
and consequently
\[
    \frac{\mathrm d}{\mathrm dt}\,2\tr(VM)
    =
    2\tr(A(N-N')).
\]

Combining the layer and cross terms gives
\begin{equation}
    \boxed{
    \Btil
    =
    P_UR_V+R_VP_U
    -
    P_VR_U-R_UP_V
    +
    2(N-N')
    }.
    \label{eq:mf_Braw_app}
\end{equation}
Thus
\[
    \dot\Psi_A
    =
    -\eta\tr(A\Btil).
\]

\paragraph{Projection onto the conservation-law space.}
Because \(A\in\Sym(r)\),
\[
    \tr(A\Btil)
    =
    \langle A,\sym(\Btil)\rangle .
\]
Hence the canonical symmetry-breaking matrix is
\[
    \Bcal
    =
    \sym(\Btil),
\]
and the matrix-valued invariant obeys
\begin{equation}
    \boxed{
    \dot\Psi
    =
    -\eta\Bcal
    =
    -\eta\sym(\Btil)
    }.
    \label{eq:mf_matrix_drift_app}
\end{equation}

The cross contribution is the expected commutator
\[
    N-N'
    =
    \E
    \left[
        [
            \delta_V^\top V,
            \delta_U^\top U
        ]
    \right].
\]
It vanishes automatically at \(r=1\), but is generally nonzero for
\(r>1\). Since a commutator need be neither symmetric nor
skew-symmetric, \(\Btil\) itself need not lie in \(\Sym(r)\); only
\(\sym(\Btil)\) is observable by the conservation laws.

At a realised time \(t\),
\[
    \dot\Psi_A(t)=0
    \quad\Longleftrightarrow\quad
    A\perp\Bcal(t),
\]
while a constant-\(A\) law survives along the whole trajectory precisely
when
\[
    A
    \perp
    \spn_t\{\Bcal(t)\}.
\]

Finally, the sufficient conditions
\[
    P_U=P_V,
    \qquad
    R_U=R_V,
    \qquad
    N=N'
\]
give \(\Btil=0\). The ordinary factor-balance condition
\(U^\top U=V^\top V\) alone does not imply the corresponding covariance
balance or vanishing of the cross term.

\begin{figure}[htpb]
    \centering
    \includegraphics[width=1.0\linewidth]{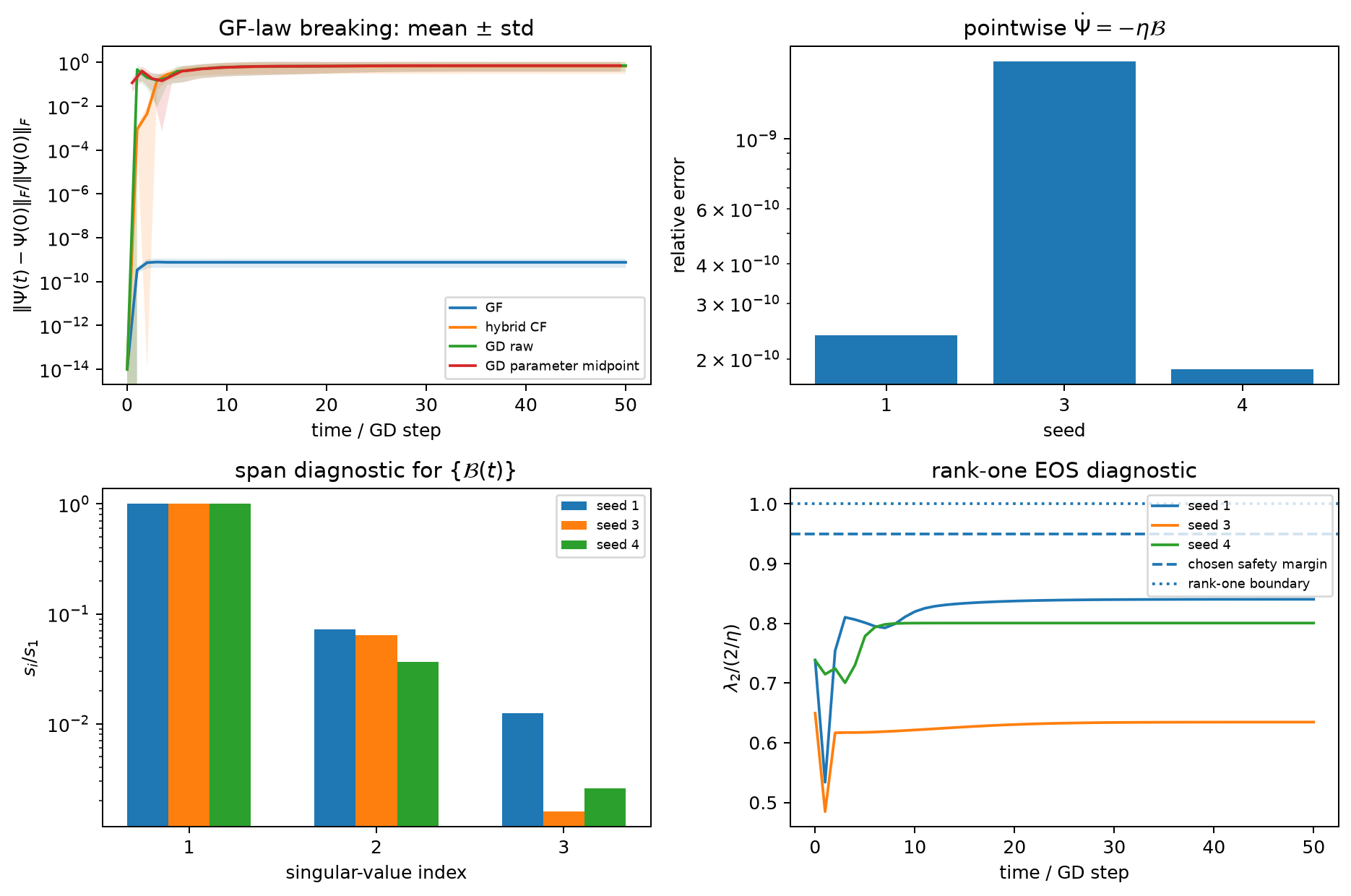}
    \caption{
\textbf{Matrix factorisation: the broken conservation-law direction over the EOS trajectory.}
We consider rank-\(r=2\) matrix factorisation, for which the gradient-flow
invariant
\(\Psi=U^\top U-V^\top V\in\Sym(2)\) contains
\(\Ngf=3\) independent scalar conservation laws. Results are shown over
three random seeds.
\textbf{Top left:} relative Frobenius displacement
\(\|\Psi(t)-\Psi(0)\|_F/\|\Psi(0)\|_F\).
Gradient flow preserves \(\Psi\) up to numerical error, whereas GD and
hybrid central flow produce an \(O(1)\) change; the adjacent-iterate GD
parameter midpoint removes the dominant period-two oscillation before
evaluating \(\Psi\).
\textbf{Top right:} pointwise validation of
\(\dot\Psi=-\eta\Bcal\), with
\(\Bcal=\sym(\Btil)\).
The relative discrepancy between the directly differentiated matrix
invariant and the symmetry-breaking prediction is approximately
\(10^{-10}\)--\(10^{-9}\) for all seeds.
\textbf{Bottom left:} normalised singular values of the matrices
\(\Bcal(t)\), represented in Frobenius-orthonormal coordinates of
\(\Sym(2)\) and stacked over the active EOS trajectory.
All three singular values are nonzero well above the numerical error floor,
showing that the realised breaking direction spans all of
\(\Sym(2)\) over the observed trajectory.
\textbf{Bottom right:} rank-one EOS diagnostic.
The second Hessian eigenvalue remains below both
\(2/\eta\) and the displayed \(0.95(2/\eta)\) safety margin throughout
the active trajectory, so the rank-one central-flow approximation remains
well separated from the onset of a second critical direction.
}
    \label{fig:mat-fac}
\end{figure}

\paragraph{Empirical analysis.}
Figure~\ref{fig:mat-fac} shows that no nonzero constant test matrix \(A\in\Sym(2)\) remains orthogonal to the breaking direction for the entire trajectory although no additional critical Hessian directions appear (the second eigenvalue stays uniformly below the EOS threshold).

\section{Two-layer ReLU}
\label{app:relu}

Consider
\[
    g(\theta,x)
    =
    U\sigma(V^\top x)
    =
    \sum_{j=1}^r
    u_j\sigma(v_j^\top x),
\]
with
\[
    L
    =
    \frac12
    \sum_i
    \|
        U\sigma(V^\top x_i)-y_i
    \|^2.
\]
The gradient-flow conservation laws are
\[
    h_j
    =
    \|u_j\|^2-\|v_j\|^2,
    \qquad
    j=1,\ldots,r,
\]
or equivalently
\[
    \Psi_A
    =
    \sum_j\lambda_j h_j,
    \qquad
    A=\operatorname{diag}(\lambda_1,\ldots,\lambda_r)
    \in\Diag(r).
\]

\paragraph{Fixed activation pattern.}
For the explicit Hessian calculation, assume that the activation pattern is
constant on the support of the oscillatory perturbations. For each example
define
\[
    z_i:=V^\top x_i,
    \qquad
    s_i:=\ind[z_i>0],
    \qquad
    D_i:=\operatorname{diag}(s_i),
\]
and
\[
    \pi_i
    :=
    \sigma(z_i)
    =
    D_i z_i,
    \qquad
    \rho_i:=U\pi_i-y_i.
\]
For \(\delta=(\delta_U,\delta_V)\), let
\[
    \alpha_i
    :=
    D_i\delta_V^\top x_i.
\]
Within this fixed-gating region,
\[
    \partial f_i(\theta)\delta 
    =
    \delta_U\pi_i+U\alpha_i,
\]
and the Hessian quadratic form is
\[
    \delta^\top H\delta
    =
    \sum_i
    \|\delta_U\pi_i+U\alpha_i\|^2
    +
    2\sum_i
    \langle\rho_i,\delta_U\alpha_i\rangle.
\]

For \(A\in\Diag(r)\), the symmetry generator satisfies
\[
    \dot U=UA,
    \qquad
    \dot V=-VA.
\]
Since \(A\) commutes with each gate matrix \(D_i\),
\[
    \dot\pi_i=-A\pi_i,
    \qquad
    \frac{\mathrm d}{\mathrm dt}(U\pi_i)
    =
    UA\pi_i-UA\pi_i
    =
    0.
\]
Thus the residual term is invariant and only the pullback term contributes.

Define
\[
    P_U:=U^\top U,
    \qquad
    R_U:=\E[\delta_U^\top\delta_U],
\]
together with
\[
    \Pi
    :=
    \sum_i\pi_i\pi_i^\top,
    \qquad
    R_V^{\rm gate}
    :=
    \sum_i\E[\alpha_i\alpha_i^\top].
\]
Expanding the pullback term gives
\[
    \sum_i
    \E[
        \|\delta_U\pi_i+U\alpha_i\|^2
    ]
    =
    \tr(\Pi R_U)
    +
    \tr(P_U R_V^{\rm gate})
    +
    C,
\]
where
\[
    C
    :=
    2\sum_i
    \E[
        \pi_i^\top
        \delta_U^\top U
        \alpha_i
    ].
\]

The Gram terms satisfy
\[
    \dot P_U=AP_U+P_UA,
    \qquad
    \dot\Pi=-(A\Pi+\Pi A),
\]
so
\[
    \frac{\mathrm d}{\mathrm dt}
    \bigl[
        \tr(P_U R_V^{\rm gate})
        +
        \tr(\Pi R_U)
    \bigr]
    =
    \tr\!\left(
        A[
            P_U R_V^{\rm gate}
            +
            R_V^{\rm gate}P_U
            -
            \Pi R_U
            -
            R_U\Pi
        ]
    \right).
\]

For the cross term, define
\[
    N_{\rm e}
    :=
    \sum_i
    \E[
        \alpha_i\pi_i^\top
        \delta_U^\top U
    ],
    \qquad
    N'_{\rm e}
    :=
    \sum_i
    \E[
        \delta_U^\top U
        \alpha_i\pi_i^\top
    ].
\]
Differentiating \(C\) and using cyclicity gives
\[
    \dot C
    =
    2\tr\!\left(
        A(N_{\rm e}-N'_{\rm e})
    \right).
\]
Hence
\begin{equation}
    \boxed{
    \Btil_{\rm ReLU}
    =
    P_U R_V^{\rm gate}
    +
    R_V^{\rm gate}P_U
    -
    \Pi R_U
    -
    R_U\Pi
    +
    2(N_{\rm e}-N'_{\rm e})
    }.
    \label{eq:relu_Braw_app}
\end{equation}
This is exactly the matrix-factorisation form under
\[
    P_V\longmapsto\Pi,
    \qquad
    R_V\longmapsto R_V^{\rm gate},
    \qquad
    N\longmapsto N_{\rm e},
    \qquad
    N'\longmapsto N'_{\rm e}.
\]

\paragraph{Projection and survival.}
For ReLU the admissible test space is \(\Diag(r)\), so
\[
    \Bcal_{\rm ReLU}
    =
    P_{\Diag(r)}
    \Btil_{\rm ReLU}.
\]
In particular,
\begin{equation}
    \boxed{
    \dot h_j
    =
    -\eta
    (\Btil_{\rm ReLU})_{jj}
    }.
    \label{eq:relu_individual_drift_app}
\end{equation}
An individual conservation law \(h_j\) therefore has zero instantaneous
drift exactly when
\[
    (\Btil_{\rm ReLU})_{jj}=0.
\]

For a general diagonal test matrix
\(A=\operatorname{diag}(\lambda_1,\ldots,\lambda_r)\),
\[
    \dot\Psi_A
    =
    -\eta
    \langle
        A,
        \Bcal_{\rm ReLU}
    \rangle .
\]
Thus individual \(h_j\)'s may all drift while a nontrivial linear
combination
\(\sum_j\lambda_jh_j\) has zero instantaneous drift.

\paragraph{Across activation boundaries.}
The matrix formula above assumes fixed gates. An exact identity for the
averaged dynamics does not require this assumption. Suppose
\[
    \dot w
    =
    -\eta\E[\nabla L(w+\delta)],
    \qquad
    \E[\delta]=0,
\]
and let the linear generator \(\Delta_A\) satisfy the symmetry identity
\[
    \langle
        \nabla L(w),
        \Delta_A(w)
    \rangle
    =
    0
\]
wherever \(L\) is differentiable. Since
\(\nabla\Psi_A=2\Delta_A\),
\[
    \dot\Psi_A
    =
    -2\eta
    \E[
        \langle
            \Delta_A(w),
            \nabla L(w+\delta)
        \rangle
    ].
\]
Using the invariance identity at \(w+\delta\) together with
\(\Delta_A(w+\delta)=\Delta_A(w)+\Delta_A(\delta)\) gives
\begin{equation}
    \boxed{
    \dot\Psi_A
    =
    2\eta
    \E
    \left[
        \left\langle
            \Delta_A(\delta),
            \nabla L(w+\delta)
        \right\rangle
    \right]
    }.
    \label{eq:relu_exact_kink_app}
\end{equation}
Identity~\eqref{eq:relu_exact_kink_app} is not specific to ReLU: it uses only
that \(\Delta_A\) is linear, that \(\nabla\Psi_A=2\Delta_A\), and that \(L\) is
invariant along \(\Delta_A\) at every point \(w+\delta\) where it is
differentiable, so it holds equally for the scalar and matrix factorisation
examples. We state it here because, for ReLU, the loss is not smooth: when \(L\)
is smooth near \(w\), the central flow \eqref{eq:central_flow} is the second-order truncation of
these averaged dynamics and \eqref{eq:relu_Braw_app} is the corresponding explicit drift, which
requires the oscillation to stay within one activation region.
Identity~\eqref{eq:relu_exact_kink_app} allows the oscillation to
cross activation boundaries between \(w-\delta\) and \(w+\delta\), only limiting the sampled points to not lie exactly on a boundary. Nevertheless, there is a dependence on the full distribution of \(\delta\), whereas the central flow only looks at \(\Sigma=\E[\delta\delta^\top]\).

\paragraph{Empirical analysis.}

\begin{figure}[htpb]
    \centering
    \includegraphics[width=1.0\linewidth]{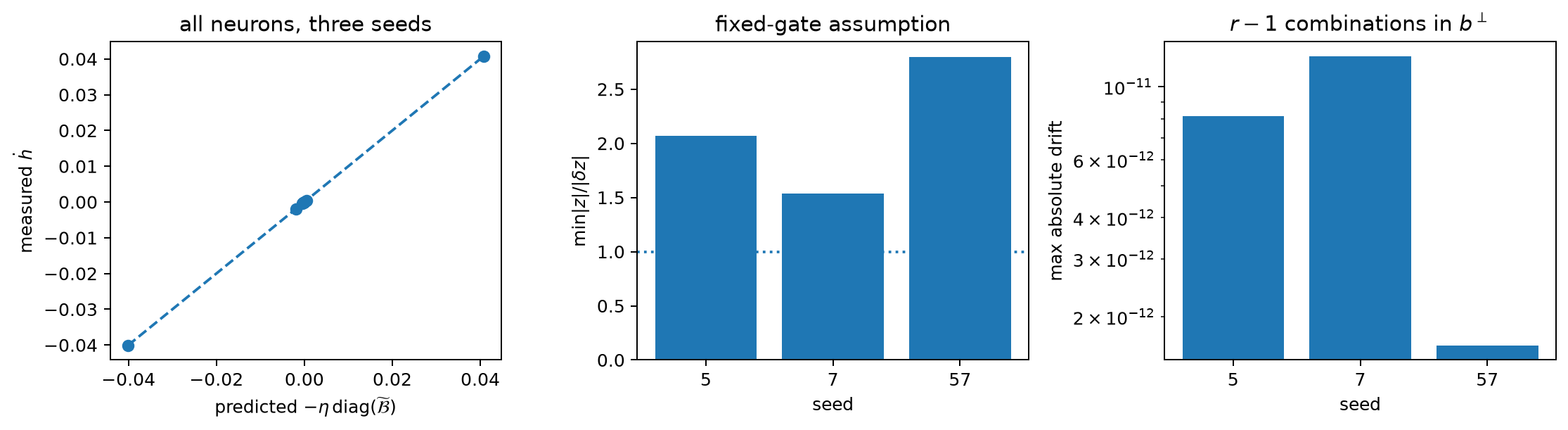}
    \caption{
\textbf{Two-layer ReLU: local symmetry breaking inside a fixed activation
region.}
We consider a two-layer ReLU network with \(r=3\) hidden units and evaluate
the central-flow prediction at an active rank-one EOS point for three
independent seeds. The admissible test space is
\(\mathcal A=\Diag(r)\), so only the diagonal projection of the raw
breaking matrix is observable.
\textbf{Left:} pointwise validation of the per-neuron drift
$
    \dot h_j
    =
    -\eta
    (\Btil_{\rm ReLU})_{jj},
    h_j=\|u_j\|^2-\|v_j\|^2.
$
Each marker corresponds to one neuron from one seed; the measured drift
\(\langle\nabla h_j,\dot\theta_{\rm CF}\rangle\) lies on the identity line
against the prediction.
\textbf{Centre:} verification of the fixed-gating assumption. For every
seed we report
$
    \min_{i,j}
    \frac{|z_{ij}|}
    {|\delta z_{ij}|},
    \delta=\pm\sqrt{\sigma^2}\,q,
$
where \(z_{ij}=v_j^\top x_i\).
All ratios exceed one, so neither side of the rank-one oscillation crosses
a ReLU activation boundary.
\textbf{Right:} instantaneous surviving combinations. Writing
\(b=\text{diag}(\Btil_{\rm ReLU})\in\R^r\), we construct an orthonormal basis of
\(b^\perp\), which has dimension \(r-1=2\) whenever \(b\neq0\).
The maximum measured absolute drift among these two combinations is
\(O(10^{-11})\) for every seed.}
    \label{fig:relu_constant_activ}
\end{figure}

Figure~\ref{fig:relu_constant_activ} verifies the architecture-dependent
projection underlying the ReLU result. The ReLU conservation laws only probe the diagonal component of the raw symmetry-breaking matrix. The directly measured \(\dot h_j\) agrees with
\(-\eta(\Btil_{\rm ReLU})_{jj}\). The agreement
includes both positive and negative drifts, showing that the calculation
predicts not only whether a law is broken but also its signed instantaneous motion. The explicit ReLU Hessian calculation is only valid while the oscillation remains inside a single activation region. 

\begin{figure}[htpb]
    \centering
    \includegraphics[width=1.0\linewidth]{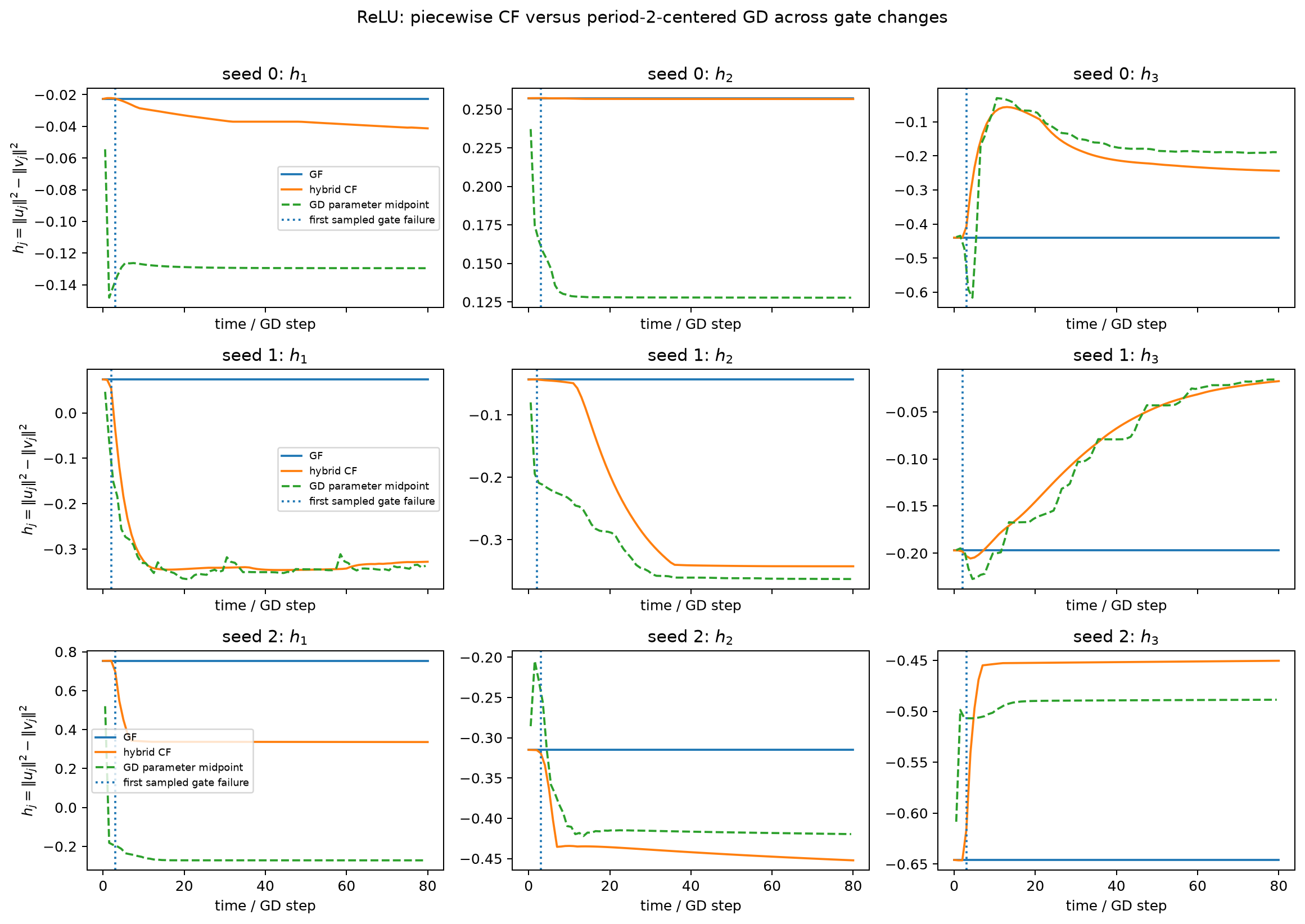}
    \caption{
\textbf{Two-layer ReLU beyond fixed gates: piecewise continuation of the central flow.}
For each of three random seeds (rows), we track the three per-neuron
gradient-flow laws
\(h_j=\|u_j\|^2-\|v_j\|^2\) (columns).
Gradient flow (solid blue) preserves every \(h_j\), while finite-step GD
breaks them. To remove the dominant period-two oscillation, the dashed
green curve evaluates \(h_j\) at the adjacent-iterate parameter midpoint
\(\overline{\theta}_{k+1/2}
=(\theta_k^{\rm GD}+\theta_{k+1}^{\rm GD})/2\).
The orange curve is the hybrid central flow, continued piecewise by
recomputing the local ReLU Hessian after activation-pattern changes.
The vertical dotted line marks the first sampled point at which the
rank-one oscillation fails the fixed-gate condition
\(
|z_{ij}|>
\sqrt{\sigma^2}|q_{v_j}^{\top}x_i|
\)
for some sample-neuron pair.
Consequently, the explicit fixed-gate symmetry-breaking formula \eqref{eq:relu_Braw_app} is
certified only to the left of this line; to the right, the orange curve is a heuristic piecewise continuation shown for illustration, with gradient flow and centred GD for reference.
}
    \label{fig:relu_across_gates}
\end{figure}

Figure~\ref{fig:relu_across_gates} shows what happens beyond the fixed-gate regime. When a pre-activation crosses zero, the Hessian changes discontinuously, and along the hybrid central-flow trajectories the rank-one oscillation violates the fixed-gate condition after fewer than 10 optimisation steps. Beyond that point ~\eqref{eq:relu_Braw_app} is no longer certified: the orange curves are obtained by recomputing the local Hessian in each new activation region. Gradient flow and centred GD are shown for reference. Characterising the central flow conservation laws across activation boundaries is left as a follow-up to this work.

\section{Experimental settings}
\label{app:settings}
The git repository can be found at: \url{https://github.com/naimaeb/CF_conserved_quantities.git} and the experimental settings are given by Table \ref{tab:settings}:
\begin{table}[h]
\centering\small
\begin{tabular}{lcccc}
\toprule
& scalar & matrix fact. & ReLU (local) & ReLU (trajectory)\\
\midrule
$\phi$ & $uv$ & $UV^{\top}$ & $U\sigma(V^{\top}x)$ & $U\sigma(V^{\top}x)$\\
dimensions & $n=m=r=1$ & $n=m=3$, $r=2$ & $n=1$, $m=4$, $r=3$ & $n=1$, $m=4$, $r=3$\\
data & $y=1$ & rank-$2$ target $Y$ & $N=24$, $x_i\sim\mathcal N(0,I)$ & as local\\
initialisation & $u_0\!\in\![1,1.45]$, $v_0\!\in\![0.05,0.3]$
               & $0.8\,\mathcal N$, $0.3\,\mathcal N$
               & $0.6\,\mathcal N$, $0.25\,\mathcal N$ & as local\\
n\_seeds & $3$ & $3$ & $3$ & $3$\\
$\eta$ & $0.90,0.95,0.96$ & $0.87,0.12,0.19$ & $2.22,0.69,2.99$ & $0.56,1.62,1.19$\\
horizon (steps) & $200$ & $50$ & --- & $80$\\
CF step & $0.02$ & $0.1$ & --- & $0.1$\\
$\Ngf$ & $1$ & $3$ & $3$ & $3$\\
\bottomrule
\end{tabular}
\caption{Configuration of the four experiments. ReLU targets are $y_i=x_i^{\top}w_\star$ with
$w_\star\sim\mathcal N(0,I)$, and the ReLU loss is a dataset mean. Note that when considering the dataset mean in the ReLU case, we need to scale the loss, the Hessian quadratic form, $\Pi$, $R_V^{\text{gate}}$, $C$, $N_\text{e}$ and $N_\text{e}'$ by $\frac{1}{N}$.}
\label{tab:settings}
\end{table}
\end{document}